\documentclass[runningheads]{llncs}
\usepackage{graphicx}
\usepackage{amsmath}
\usepackage{amssymb}
\usepackage{booktabs}
\usepackage{multirow}
\usepackage{hyperref}
\usepackage[misc]{ifsym}

\begin{document}
\title{AME-CAM: Attentive Multiple-Exit CAM for Weakly Supervised Segmentation on MRI Brain Tumor}
% \title{AME-CAM: A Novel Weakly Supervised MRI Brain Tumor Segmentation Method}

\author{Yu-Jen Chen\inst{1} %index{Chen, Yu-Jen}
\and Xinrong Hu\inst{2} %index{Hu, Xinrong}
\and Yiyu Shi\inst{2} %index{Shi, Yiyu}
\and Tsung-Yi Ho\inst{3} %index{Ho, Tsung-Yi}
}

\authorrunning{Y.-J. Chen et al.}

\institute{National Tsing Hua University, Taiwan\\ 
\email{yujenchen@gapp.nthu.edu.tw}
\and University of Notre Dame, Notre Dame, IN, USA\\
\email{\{xhu7, yshi4\}@nd.edu}
\and
The Chinese University of Hong Kong, Hong Kong\\
\email{tyho@cse.cuhk.edu.hk}}
\titlerunning{AME-CAM for WSSS}
\maketitle              % typeset the header of the contribution
\begin{abstract}

Magnetic resonance imaging (MRI) is commonly used for brain tumor segmentation, which is critical for patient evaluation and treatment planning. To reduce the labor and expertise required for labeling, weakly-supervised semantic segmentation (WSSS) methods with class activation mapping (CAM) have been proposed. However, existing CAM methods suffer from low resolution due to strided convolution and pooling layers, resulting in inaccurate predictions. In this study, we propose a novel CAM method, Attentive Multiple-Exit CAM (AME-CAM), that extracts activation maps from multiple resolutions to hierarchically aggregate and improve prediction accuracy. We evaluate our method on the BraTS 2021 dataset and show that it outperforms state-of-the-art methods.
\end{abstract}

\section{Introduction}
\label{sec:intro}

Deep learning techniques have greatly improved medical image segmentation by automatically extracting specific tissue or substance location information, which facilitates accurate disease diagnosis and assessment. However, most deep learning approaches for segmentation require fully or partially labeled training datasets, which can be time-consuming and expensive to annotate. To address this issue, recent research has focused on developing segmentation frameworks that require little or no segmentation labels.

To meet this need, many researchers have devoted their efforts to Weakly-Supervised Semantic Segmentation (WSSS)\cite{wolleb2022diffusion}, which utilizes weak supervision, such as image-level classification labels. Recent WSSS methods can be broadly categorized into two types \cite{chan2021comprehensive}: Class-Activation-Mapping-based (CAM-based) \cite{selvaraju2017grad,wang2020score,jiang2021layercam,wang2020self,lee2021lfi,xie2022c2am}, and Multiple-Instance-Learning-based (MIL-based) \cite{qian2022transformer} methods.

The literature has not adequately addressed the issue of low-resolution Class-Activation Maps (CAMs), especially for medical images. Some existing methods, such as dilated residual networks \cite{yu2017dilated} and U-Net segmentation architecture \cite{belharbi2022f,englebert2022poly,tagaris2019high}, have attempted to tackle this issue, but still require many upsampling operations, which the results become blurry. Meanwhile, LayerCAM \cite{jiang2021layercam} has proposed a hierarchical solution that extracts activation maps from multiple convolution layers using Grad-CAM\cite{selvaraju2017grad} and aggregates them with equal weights. Although this approach successfully enhances the resolution of the segmentation mask, it lacks flexibility and may not be optimal.

In this paper, we propose an Attentive Multiple-Exit CAM (AME-CAM) for brain tumor segmentation in magnetic resonance imaging (MRI). Different from recent CAM methods, AME-CAM uses a classification model with multiple-exit training strategy applied to optimize the internal outputs. Activation maps from the outputs of internal classifiers, which have different resolutions, are then aggregated using an attention model. The model learns the pixel-wise weighted sum of the activation maps by a novel contrastive learning method.

Our proposed method has the following contributions:
\begin{itemize}
\item To tackle the issues in existing CAMs, we propose to use multiple-exit classification networks to accurately capture all the internal activation maps of different resolutions.
\item We propose an attentive feature aggregation to learn the pixel-wise weighted sum of the internal activation maps.
\item We demonstrate the superiority of AME-CAM over state-of-the-art CAM methods in extracting segmentation results from classification networks on the 2021 Brain Tumor Segmentation Challenge (BraTS 2021) \cite{menze2014multimodal,bakas2017advancing,bakas2018identifying}.
\item For reproducibility, we have released our code at \\ \href{https://github.com/windstormer/AME-CAM}{https://github.com/windstormer/AME-CAM}
\end{itemize}

Overall, our proposed method can help overcome the challenges of expensive and time-consuming segmentation labeling in medical imaging, and has the potential to improve the accuracy of disease diagnosis and assessment.

% \section{Motivation and Related Works}
% \label{sec:related_works}
% \input{Section/Related_Works.tex}

\section{Attentive Multiple-Exit CAM (AME-CAM)}
\label{sec:methodology}

\begin{figure}[ht]
\begin{center}
\includegraphics[width=0.9\linewidth]{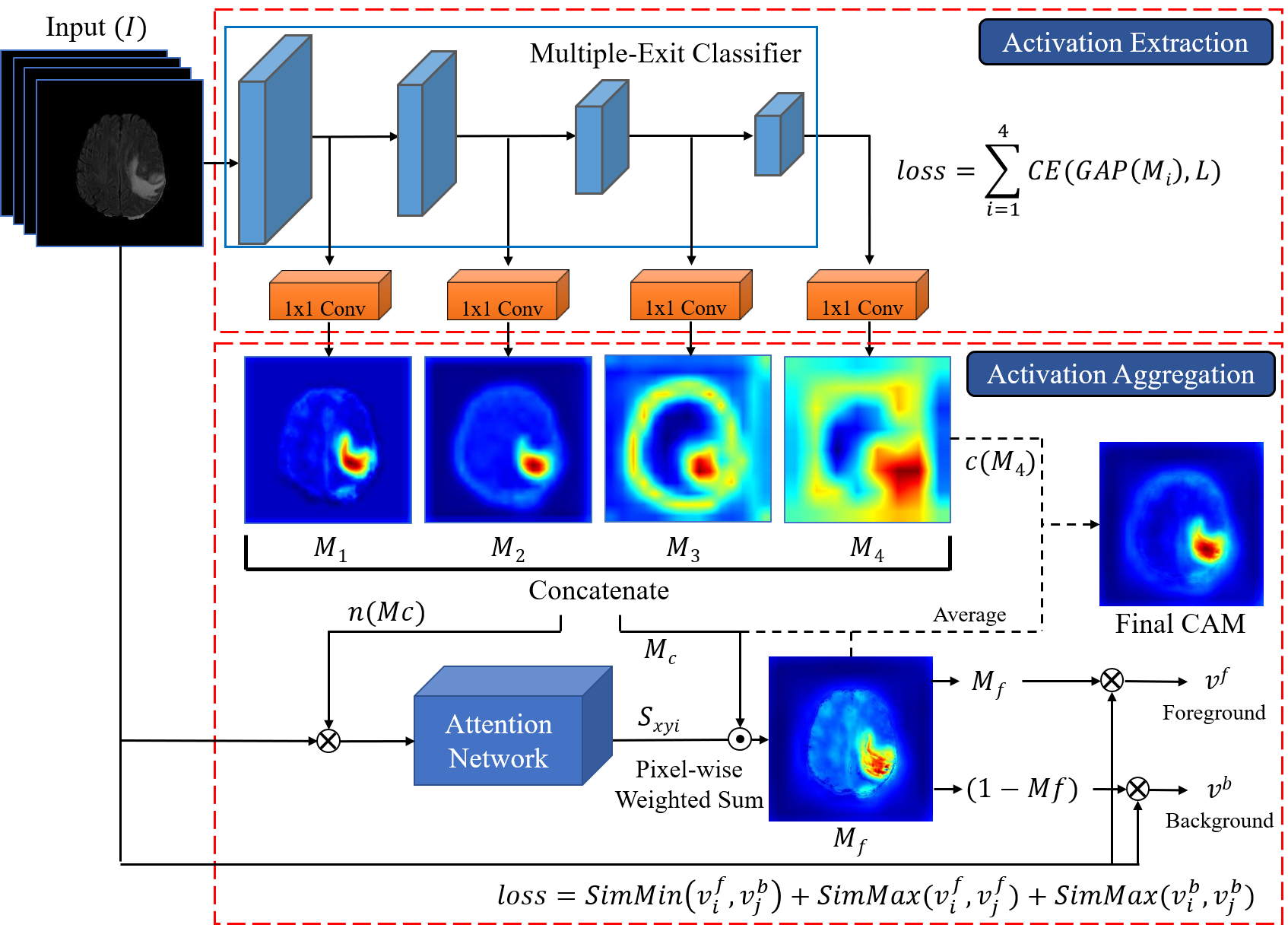}

\caption{An overview of the proposed AME-CAM method, which contains multiple-exit network based activation extraction phase and attention based activation aggregation phase. The operator $\odot$ and $\otimes$ denote the pixel-wise weighted sum and the pixel-wise multiplication, respectively.
}
\label{Fig.AME-CAM}
\end{center}
\end{figure}

The proposed AME-CAM method consists of two training phases: activation extraction and activation aggregation, as shown in Fig. \ref{Fig.AME-CAM}. In the activation extraction phase, we use a binary classification network, e.g., ResNet-18, to obtain the class probability $y=f(I)$ of the input image $I$. To enable multiple-exit training, we add one internal classifier after each residual block, which generates the activation map $M_i$ of different resolutions. We use a cross-entropy loss to train the multiple-exit classifier, which is defined as
\begin{equation}
\label{eq.mloss}
loss = \sum_{i=1}^4 CE(GAP(M_i), L)
\end{equation}
where $GAP(\cdot)$ is the global-average-pooling operation, $CE(\cdot)$ is the cross-entropy loss, and $L$ is the image-wise ground-truth label.

In the activation aggregation phase, we create an efficient hierarchical aggregation method to generate the aggregated activation map $M_f$ by calculating the pixel-wise weighted sum of the activation maps $M_i$. We use an attention network $A(\cdot)$ to estimate the importance of each pixel from each activation map. The attention network takes in the input image $I$ masked by the activation map and outputs the pixel-wised importance score $S_{xyi}$ of each activation map. We formulate the operation as follows:
\begin{equation}
\label{eq.anet}
S_{xyi} = A([I\otimes n(M_i)]_{i=1}^4)
\end{equation}
where $[\cdot]$ is the concatenate operation, $n(\cdot)$ is the min-max normalization to map the range to [0,1], and $\otimes$ is the pixel-wise multiplication, which is known as image masking. The aggregated activation map $M_f$ is then obtained by the pixel-wise weighted sum of $M_i$, which is $M_f = \sum_{i=1}^4 (S_{xyi} \otimes M_i)$.

We train the attention network with unsupervised contrastive learning, which forces the network to disentangle the foreground and the background of the aggregated activation map $M_f$. We mask the input image by the aggregated activation map $M_f$ and its opposite $(1-M_f)$ to obtain the foreground feature and the background feature, respectively. The loss function is defined as follows:
\begin{equation}
\label{eq.closs}
\begin{aligned}
loss = SimMin(v_i^f, v_j^b)+ SimMax(v_i^f, v_j^f) + SimMax(v_i^b, v_j^b)
\end{aligned}
\end{equation}
where $v_i^f$ and $v_i^b$ denote the foreground and the background feature of the i-th sample, respectively. $SimMin$ and $SimMax$ are the losses that minimize and maximize the similarity between two features (see C$^2$AM\cite{xie2022c2am} for details).

Finally, we average the activation maps $M_1$ to $M_4$ and the aggregated map $M_f$ to obtain the final CAM results for each image. We apply the Dense Conditional Random Field (DenseCRF)\cite{krahenbuhl2011efficient} algorithm to generate the final segmentation mask. It is worth noting that the proposed method is flexible and can be applied to any classification network architecture.

\begin{table}
\centering
\caption{Comparison with weakly supervised methods (WSSS), unsupervised method (UL), and fully supervised methods (FSL) on BraTS dataset with T1, T1-CE, T2, and T2-FLAIR MRI images. Results are reported in the form of mean$\pm$std. We mark the highest score among WSSS methods with bold text.}
\label{Table:Compare_CAM_BraTS}
\begin{tabular}{c|c|c|c|l}
\hline
\multicolumn{5}{c}{BraTS T1} \\ \hline
Type           & Method            & Dice $\uparrow$         & IoU $\uparrow$          & \multicolumn{1}{c}{HD95 $\downarrow$}            \\ \hline
\multirow{6}{*}{WSSS}           & Grad-CAM (2016)   & 0.107$\pm$0.090 & 0.059$\pm$0.055 & 121.816$\pm$22.963 \\
               & ScoreCAM (2020)   & 0.296$\pm$0.128 & 0.181$\pm$0.089 & 60.302$\pm$14.110  \\
               & LFI-CAM (2021)    & 0.568$\pm$0.167 & 0.414$\pm$0.152 & 23.939$\pm$25.609  \\
               & LayerCAM (2021)   & 0.571$\pm$0.170 & 0.419$\pm$0.161 & 23.335$\pm$27.369  \\
               & Swin-MIL (2022)   & 0.477$\pm$0.170 & 0.330$\pm$0.147 & 46.468$\pm$30.408  \\
               & AME-CAM  (ours)   & \textbf{0.631$\pm$0.119} & \textbf{0.471$\pm$0.119} & \textbf{21.813$\pm$18.219}  \\ \hline
UL             & C\&F (2020)       & 0.200$\pm$0.082 & 0.113$\pm$0.051 & 79.187$\pm$14.304  \\ \hline
\multirow{2}{*}{FSL}            & C\&F (2020)       & 0.572$\pm$0.196 & 0.426$\pm$0.187 & 29.027$\pm$20.881  \\
               & Opt. U-net (2021) & 0.836$\pm$0.062 & 0.723$\pm$0.090 & 11.730$\pm$10.345  \\ \hline
\multicolumn{5}{c}{BraTS T1-CE} \\ \hline
Type           & Method            & Dice $\uparrow$         & IoU $\uparrow$          & \multicolumn{1}{c}{HD95 $\downarrow$}            \\ \hline
\multirow{6}{*}{WSSS}           & Grad-CAM (2016)   & 0.127$\pm$0.088 & 0.071$\pm$0.054 & 129.890$\pm$27.854 \\
               & ScoreCAM (2020)   & 0.397$\pm$0.189 & 0.267$\pm$0.163 & 46.834$\pm$22.093  \\
               & LFI-CAM (2021)    & 0.121$\pm$0.120 & 0.069$\pm$0.076 & 136.246$\pm$38.619 \\
               & LayerCAM (2021)   & 0.510$\pm$0.209 & 0.367$\pm$0.180 & 29.850$\pm$45.877  \\
               & Swin-MIL (2022)   & 0.460$\pm$0.169 & 0.314$\pm$0.140 & 46.996$\pm$22.821  \\
               & AME-CAM  (ours)   & \textbf{0.695$\pm$0.095} & \textbf{0.540$\pm$0.108} & \textbf{18.129$\pm$12.335}  \\ \hline
UL             & C\&F (2020)       & 0.179$\pm$0.080 & 0.101$\pm$0.050 & 77.982$\pm$14.042  \\ \hline
\multirow{2}{*}{FSL}            & C\&F (2020)       & 0.246$\pm$0.104 & 0.144$\pm$0.070 & 130.616$\pm$9.879  \\
               & Opt. U-net (2021) & 0.845$\pm$0.058 & 0.736$\pm$0.085 & 11.593$\pm$11.120  \\ \hline
\multicolumn{5}{c}{BraTS T2} \\ \hline
Type           & Method            & Dice $\uparrow$         & IoU $\uparrow$          & \multicolumn{1}{c}{HD95 $\downarrow$}            \\ \hline
\multirow{6}{*}{WSSS}           & Grad-CAM (2016)   & 0.049$\pm$0.058 & 0.026$\pm$0.034 & 141.025$\pm$23.107 \\
               & ScoreCAM (2020)   & 0.530$\pm$0.184 & 0.382$\pm$0.174 & 28.611$\pm$11.596  \\
               & LFI-CAM (2021)    & 0.673$\pm$0.173 & 0.531$\pm$0.186 & 18.165$\pm$10.475  \\
               & LayerCAM (2021)   & 0.624$\pm$0.178 & 0.476$\pm$0.173 & 23.978$\pm$44.323  \\
               & Swin-MIL (2022)   & 0.437$\pm$0.149 & 0.290$\pm$0.117 & 38.006$\pm$30.000  \\
               & AME-CAM  (ours)   & \textbf{0.721$\pm$0.086} & \textbf{0.571$\pm$0.101} & \textbf{14.940$\pm$8.736}   \\ \hline
UL             & C\&F (2020)       & 0.230$\pm$0.089 & 0.133$\pm$0.058 & 76.256$\pm$13.192  \\ \hline
\multirow{2}{*}{FSL}            & C\&F (2020)       & 0.611$\pm$0.221 & 0.474$\pm$0.217 & 109.817$\pm$27.735 \\
               & Opt. U-net (2021) & 0.884$\pm$0.064 & 0.798$\pm$0.098 & 8.349$\pm$9.125    \\ \hline
\multicolumn{5}{c}{BraTS T2-FLAIR} \\ \hline
Type           & Method            & Dice $\uparrow$         & IoU $\uparrow$          & \multicolumn{1}{c}{HD95 $\downarrow$}            \\ \hline
\multirow{6}{*}{WSSS}           & Grad-CAM (2016)   & 0.150$\pm$0.077 & 0.083$\pm$0.050 & 110.031$\pm$23.307 \\
               & ScoreCAM (2020)   & 0.432$\pm$0.209 & 0.299$\pm$0.178 & 39.385$\pm$17.182  \\
               & LFI-CAM (2021)    & 0.161$\pm$0.192 & 0.102$\pm$0.140 & 125.749$\pm$45.582 \\
               & LayerCAM (2021)   & 0.652$\pm$0.206 & 0.515$\pm$0.210 & 22.055$\pm$33.959  \\
               & Swin-MIL (2022)   & 0.272$\pm$0.115 & 0.163$\pm$0.079 & 41.870$\pm$19.231  \\
               & AME-CAM  (ours)   & \textbf{0.862$\pm$0.088} & \textbf{0.767$\pm$0.122} & \textbf{8.664$\pm$6.440}    \\ \hline
UL             & C\&F (2020)       & 0.306$\pm$0.190 & 0.199$\pm$0.167 & 75.651$\pm$14.214  \\ \hline
\multirow{2}{*}{FSL}            & C\&F (2020)       & 0.578$\pm$0.137 & 0.419$\pm$0.130 & 138.138$\pm$14.283 \\
               & Opt. U-net (2021) & 0.914$\pm$0.058 & 0.847$\pm$0.093 & 8.093$\pm$11.879   \\ \hline
\end{tabular}
\end{table}

\section{Experiments}
\label{sec:experiments}

\subsection{Dataset}

We evaluate our method on the Brain Tumor Segmentation challenge (BraTS) dataset \cite{menze2014multimodal,bakas2017advancing,bakas2018identifying}, which contains 2,000 cases, each of which includes four 3D volumes from four different MRI modalities: T1, post-contrast enhanced T1 (T1-CE), T2, and T2 Fluid Attenuated Inversion Recovery (T2-FLAIR), as well as a corresponding segmentation ground-truth mask. The official data split divides these cases by the ratio of 8:1:1 for training, validation, and testing (5,802 positive and 1,073 negative images). In order to evaluate the performance, we use the validation set as our test set and report statistics on it. We preprocess the data by slicing each volume along the z-axis to form a total of 193,905 2D images, following the approach of Kang et al. \cite{kang2021towards} and Dey and Hong \cite{dey2021asc}. We use the ground-truth segmentation masks only in the final evaluation, not in the training process.

\subsection{Implementation Details and Evaluation Protocol}

We implement our method in PyTorch using ResNet-18 as the backbone classifier. We pretrain the classifier using SupCon \cite{khosla2020supervised} and then fine-tune it in our experiments. We use the entire training set for both pretraining and fine-tuning. We set the initial learning rate to 1e-4 for both phases, and use the cosine annealing scheduler to decrease it until the minimum learning rate is 5e-6. We set the weight decay in both phases to 1e-5 for model regularization. We use Adam optimizer in the multiple-exit phase and SGD optimizer in the aggregation phase. We train all classifiers until they converge with a test accuracy of over 0.9 for all image modalities. Note that only class labels are available in the training set.

We use the Dice score and Intersection over Union (IoU) to evaluate the quality of the semantic segmentation, following the approach of Xu et al. \cite{xu2019whole}, Tang et al. \cite{tang2021m}, and Qian et al. \cite{qian2022transformer}. In addition, we report the 95\% Hausdorff Distance (HD95) to evaluate the boundary of the prediction mask.

Interested readers can refer to the supplementary material for results on other network architectures.

\section{Results}
\label{sec:results}

\begin{figure}[ht]
\begin{center}
\includegraphics[width=0.95\linewidth]{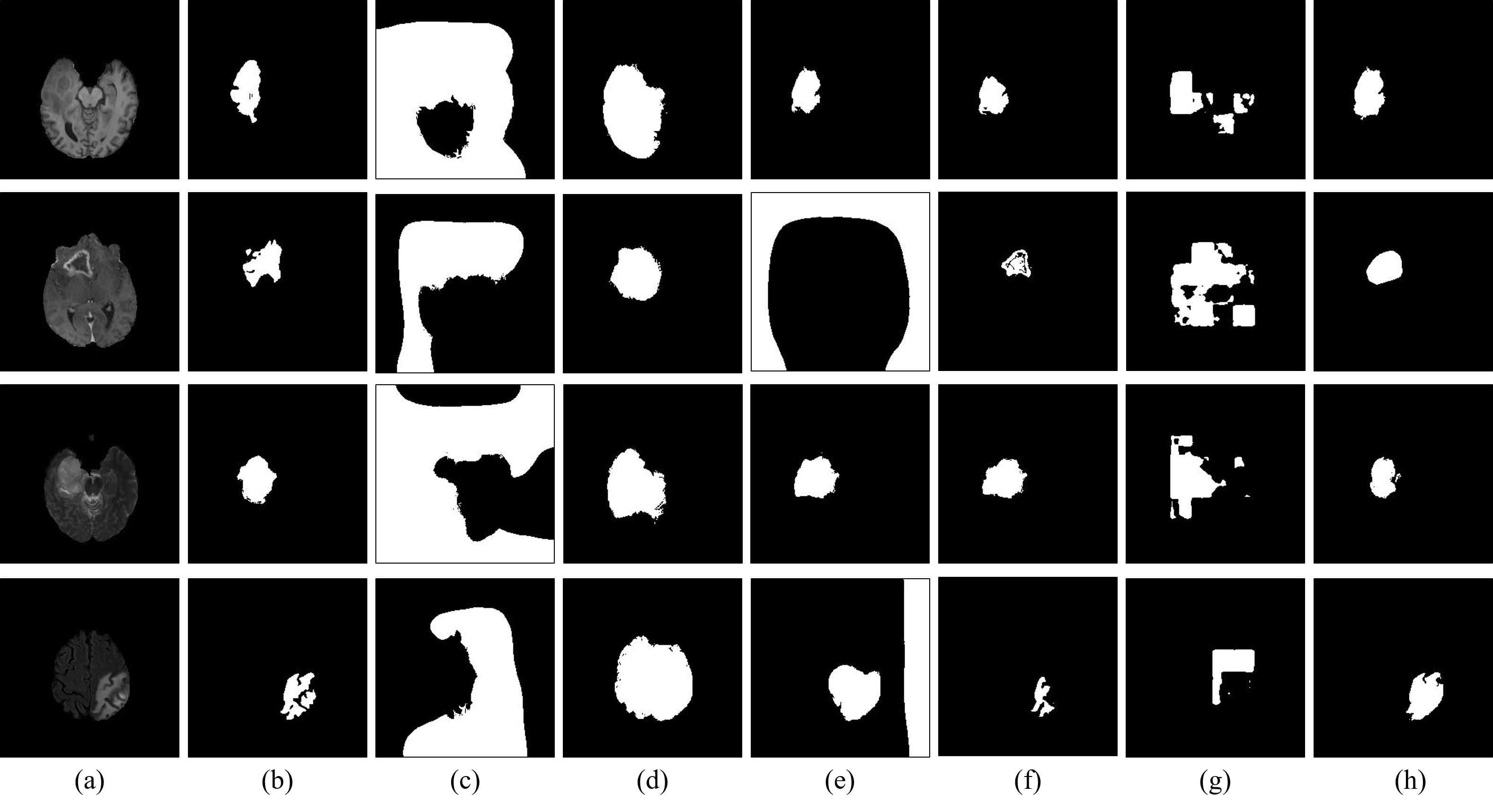}

\caption{Qualitative results of all methods. (a) Input Image. (b) Ground Truth. (c) Grad-CAM \cite{selvaraju2017grad} (d) ScoreCAM \cite{wang2020score}. (e) LFI-CAM \cite{lee2021lfi}. (f) LayerCAM \cite{jiang2021layercam}. (g) Swin-MIL \cite{qian2022transformer}. (h) AME-CAM (ours). The image modalities of rows 1-4 are T1, T1-CE, T2, T2-FLAIR, respectively from the BraTS dataset.
}
\label{Fig.Result}
\end{center}
\end{figure}

\subsection{Quantitative and Qualitative Comparison with State-of-the-art}
\label{result:cam_sota}

In this section, we compare the segmentation performance of the proposed AME-CAM with five state-of-the-art weakly-supervised segmentation methods, namely Grad-CAM \cite{selvaraju2017grad}, ScoreCAM \cite{wang2020score}, LFI-CAM \cite{lee2021lfi}, LayerCAM \cite{jiang2021layercam}, and Swin-MIL \cite{qian2022transformer}. We also compare with an unsupervised approach C\&F \cite{chen2020medical}, the supervised version of C\&F, and the supervised Optimized U-net \cite{futrega2021optimized} to show the comparison with non-CAM-based methods. We acknowledge that the results from fully supervised and unsupervised methods are not directly comparable to the weakly supervised CAM methods. Nonetheless, these methods serve as interesting references for the potential performance ceiling and floor of all the CAM methods.

Quantitatively, Grad-CAM and ScoreCAM result in low dice scores, demonstrating that they have difficulty extracting the activation of medical images. LFI-CAM and LayerCAM improve the dice score in all modalities, except LFI-CAM in T1-CE and T2-FLAIR. Finally, the proposed AME-CAM achieves optimal performance in all modalities of the BraTS dataset.

Compared to the unsupervised baseline (UL), C\&F is unable to separate the tumor and the surrounding tissue due to low contrast, resulting in low dice scores in all experiments. With pixel-wise labels, the dice of supervised C\&F improves significantly. Without any pixel-wise label, the proposed AME-CAM outperforms supervised C\&F in all modalities.

The fully supervised (FSL) Optimized U-net achieves the highest dice score and IoU score in all experiments. However, even under different levels of supervision, there is still a performance gap between the weakly supervised CAM methods and the fully supervised state-of-the-art. This indicates that there is still potential room for WSSS methods to improve in the future.

Qualitatively, Fig. \ref{Fig.Result} shows the visualization of the CAM and segmentation results from all six CAM-based approaches under four different modalities from the BraTS dataset. Grad-CAM (Fig. \ref{Fig.Result}(c)) results in large false activation region, where the segmentation mask is totally meaningless. ScoreCAM eliminates false activation corresponding to air. LFI-CAM focus on the exact tumor area only in the T1 and T2 MRI (row 1 and 3). Swin-MIL can hardly capture the tumor region of the MRI image, where the activation is noisy. Among all, only LayerCAM and the proposed AME-CAM successfully focus on the exact tumor area, but AME-CAM reduces the under-estimation of the tumor area. This is attributed to the benefit provided by aggregating activation maps from different resolutions.

\subsection{Ablation Study}

\begin{table}
\centering
\caption{Ablation study for aggregation phase using T1 MRI images from the BraTS dataset. Avg. ME denotes that we directly average four activation maps generated by the multiple-exit phase. The dice score, IoU, and the HD95 are reported in the form of mean$\pm$std.}
\label{Table:ablation_aggregation}
\small
\begin{tabular}{c|c|c|l}
\hline
Method                            & Dice $\uparrow$        & IoU $\uparrow$         & \multicolumn{1}{c}{HD95 $\downarrow$}           \\ \hline
Avg. ME                           & 0.617$\pm$0.121 & 0.457$\pm$0.121 & 23.603$\pm$20.572  \\
Avg. ME+C$^2$AM\cite{xie2022c2am} & 0.484$\pm$0.256 & 0.354$\pm$0.207 & 69.242$\pm$121.163 \\
AME-CAM (ours)                    & \textbf{0.631$\pm$0.119} & \textbf{0.471$\pm$0.119} & \textbf{21.813$\pm$18.219} \\ \hline
\end{tabular}
\end{table}
\textbf{Effect of Different Aggregation Approaches:} In Table \ref{Table:ablation_aggregation}, we conducted an ablation study to investigate the impact of using different aggregation approaches after extracting activations from the multiple-exit network. We aim to demonstrate the superiority of the proposed attention-based aggregation approach for segmenting tumor regions in T1 MRI of the BraTS dataset. Note that we only report the results for T1 MRI in the BraTS dataset. Please refer to the supplementary material for the full set of experiments.

As a baseline, we first conducted the average of four activation maps generated by the multiple-level activation extraction (Avg. ME). We then applied C$^2$AM \cite{xie2022c2am}, a state-of-the-art CAM-based refinement approach, to refine the result of the baseline, which we call "Avg. ME+C$^2$AM". However, we observed that C$^2$AM tended to segment the brain region instead of the tumor region due to the larger contrast between the brain tissue and the air than that between the tumor region and its surrounding tissue. Any incorrect activation of C$^2$AM also led to inferior results, resulting in a degradation of the average dice score from 0.617 to 0.484. In contrast, the proposed attention-based approach provided a significant weighting solution that led to optimal performance in all cases.

\begin{table}[h]
\centering
\caption{Ablation study for using single-exit from $M_1$, $M_2$, $M_3$ or $M_4$ of Fig. \ref{Fig.AME-CAM} and the multiple-exit using results from $M_2$ and $M_3$ and using all exits (AME-CAM). The experiments are done on the T1-CE MRI of BraTS dataset. The dice score, IoU, and the HD95 are reported in the form of mean$\pm$std.}
\label{Table:ablation_single_exit}
\begin{tabular}{c|c|c|c|l}
\hline
\multicolumn{2}{c|}{Selected Exit}                & Dice $\uparrow$        & IoU $\uparrow$         & \multicolumn{1}{c}{HD95 $\downarrow$}           \\ \hline
\multirow{4}{*}{Single-exit}   & $M_1$          & 0.144$\pm$0.184 & 0.090$\pm$0.130 & 74.249$\pm$62.669           \\
              & $M_2$          & 0.500$\pm$0.231 & 0.363$\pm$0.196 & 43.762$\pm$85.703           \\
              & $M_3$          & 0.520$\pm$0.163 & 0.367$\pm$0.141 & 43.749$\pm$54.907           \\
              & $M_4$          & 0.154$\pm$0.101 & 0.087$\pm$0.065 & 120.779$\pm$44.548          \\ \hline
\multirow{2}{*}{Multiple-exit} & $M_2+M_3$      & 0.566$\pm$0.207 & 0.421$\pm$0.186 & 27.972$\pm$56.591           \\
              & AME-CAM (ours) & \textbf{0.695$\pm$0.095} & \textbf{0.540$\pm$0.108} & \textbf{18.129$\pm$12.335}           \\ \hline
\end{tabular}
\end{table}
\textbf{Effect of Single-Exit and Multiple-Exit:} Table \ref{Table:ablation_single_exit} summarizes the performance of using single-exit from $M_1$, $M_2$, $M_3$, or $M_4$ of Fig. \ref{Fig.AME-CAM} and the multiple-exit using results from $M_2$ and $M_3$, and using all exits (AME-CAM) on T1-CE MRI in the BraTS dataset.

The comparisons show that the activation map obtained from the shallow layer $M_1$ and the deepest layer $M_4$ result in low dice scores, around 0.15. This is because the network is not deep enough to learn the tumor region in the shallow layer, and the resolution of the activation map obtained from the deepest layer is too low to contain sufficient information to make a clear boundary for the tumor. Results of the internal classifiers from the middle of the network ($M_2$ and $M_3$) achieve the highest dice score and IoU, both of which are around 0.5.

To evaluate whether using results from all internal classifiers leads to the highest performance, we further apply the proposed method to the two internal classifiers with the highest dice scores, i.e., $M_2$ and $M_3$, called $M_2+M_3$. Compared with using all internal classifiers ($M_1$ to $M_4$), $M_2+M_3$ results in 18.6\% and 22.1\% lower dice and IoU, respectively. In conclusion, our AME-CAM still achieves the optimal performance among all the experiments of single-exit and multiple-exit.

Other ablation studies are presented in the supplementary material due to space limitations.

\section{Conclusion}
\label{sec:conclusion}

In this work, we propose a brain tumor segmentation method for MRI images using only class labels, based on an Attentive Multiple-Exit Class Activation Mapping (AME-CAM). Our approach extracts activation maps from different exits of the network to capture information from multiple resolutions. We then use an attention model to hierarchically aggregate these activation maps, learning pixel-wise weighted sums.

Experimental results on the four modalities of the 2021 BraTS dataset demonstrate the superiority of our approach compared with other CAM-based weakly-supervised segmentation methods. Specifically, AME-CAM achieves the highest dice score for all patients in all datasets and modalities. These results indicate the effectiveness of our proposed approach in accurately segmenting brain tumors from MRI images using only class labels.

{\small
\bibliographystyle{splncs04}
\bibliography{miccai23}
}

\end{document}

%
% \title{AME-CAM: Attentive Multiple-Exit CAM for Weakly Supervised Segmentation on MRI Brain Tumor}
%
%\titlerunning{Abbreviated paper title}
% If the paper title is too long for the running head, you can set
% an abbreviated paper title here
%
% \author{First Author\inst{1}\orcidID{0000-1111-2222-3333} \and
% Second Author\inst{2,3}\orcidID{1111-2222-3333-4444} \and
% Third Author\inst{3}\orcidID{2222--3333-4444-5555}}
% %
% \authorrunning{F. Author et al.}
% % First names are abbreviated in the running head.
% % If there are more than two authors, 'et al.' is used.
% %
% \institute{Princeton University, Princeton NJ 08544, USA \and
% Springer Heidelberg, Tiergartenstr. 17, 69121 Heidelberg, Germany
% \email{lncs@springer.com}\\
% \url{http://www.springer.com/gp/computer-science/lncs} \and
% ABC Institute, Rupert-Karls-University Heidelberg, Heidelberg, Germany\\
% \email{\{abc,lncs\}@uni-heidelberg.de}}
%
% \maketitle              % typeset the header of the contribution

% \section{Comparison on the TCGA-LGG dataset}
\begin{table}
\centering
\caption{Comparison with weakly supervised methods (WSSS), unsupervised method (UL), and fully supervised methods (FSL) on The Cancer Genome Atlas Lower-Grade Glioma collection (TCGA-LGG) dataset. Results are reported in the form of mean$\pm$std. We mark the highest score among WSSS methods with bold text.}
\label{Table:Supple_TCGA-LGG}
\begin{tabular}{c|c|c|c|l}
\hline
Type                  & Method            & Dice $\uparrow$       & IoU $\uparrow$        & \multicolumn{1}{c}{HD95 $\downarrow$} \\ \hline
\multirow{6}{*}{WSSS} & Grad-CAM (2016)   & 0.407$\pm$0.229          & 0.283$\pm$0.190          & 81.278$\pm$38.648                        \\
                      & ScoreCAM (2020)   & 0.490$\pm$0.222          & 0.351$\pm$0.184          & 57.821$\pm$30.275                        \\
                      & LFI-CAM (2021)    & 0.451$\pm$0.242          & 0.323$\pm$0.206          & 62.077$\pm$36.067                        \\
                      & LayerCAM (2021)   & 0.378$\pm$0.211          & 0.255$\pm$0.171          & 69.696$\pm$42.493                        \\
                      & Swin-MIL (2022)   & 0.271$\pm$0.144          & 0.165$\pm$0.101          & 111.634$\pm$17.670                       \\
                      & AME-CAM (ours)    & \textbf{0.517$\pm$0.244} & \textbf{0.385$\pm$0.227} & \textbf{64.766$\pm$35.706}               \\ \hline
UL                    & C\&F (2020)       & 0.300$\pm$0.162          & 0.188$\pm$0.119          & 101.980$\pm$18.358                       \\ \hline
\multirow{2}{*}{FSL}  & C\&F (2020)       & 0.637$\pm$0.216          & 0.502$\pm$0.224          & 46.925$\pm$17.889                        \\
                      & Opt. U-Net (2021) & 0.668$\pm$0.196          & 0.532$\pm$0.209          & 47.035$\pm$19.529                        \\ \hline 
\end{tabular}
\end{table}

% The Cancer Genome Atlas Lower-Grade Glioma collection (TCGA-LGG) dataset \cite{pedano2016radiology} consists of 110 patients with 3,600 2D images collected from the TCGA lower-grade glioma collection who had preoperative imaging data available, containing at least a fluid-attenuated inversion recovery (FLAIR) sequence. We follow the preprocess step and the split rate in the BraTS dataset to separate the dataset into training, validation, and testing (180 positive and 351 negative images) for experiments.

% \section{Ablation study of Using Different Loss Function in Activation Aggregation}
\begin{table}[h]
\centering
\caption{Ablation study for using different loss function in the aggregation phase, conventional cross-entropy loss v.s. the proposed C$^2$AM loss. The experiments are done with the BraTS dataset. Results are reported in the form of mean$\pm$std. We mark the highest score among WSSS methods with bold text.}
\label{Table:Supple_agg_phase_loss}
\begin{tabular}{c|c|c|c|l}
\hline
Dataset                                       & Method         & Dice $\uparrow$       & IoU $\uparrow$        & \multicolumn{1}{c}{HD95 $\downarrow$} \\ \hline
\multirow{2}{*}{\shortstack{BraTS\\T1}}       & Cross-Entropy  & 0.494$\pm$0.174          & 0.345$\pm$0.154          & 43.903$\pm$34.428                        \\
                                              & C$^2$AM (ours) & \textbf{0.631$\pm$0.119} & \textbf{0.471$\pm$0.119} & \textbf{21.813$\pm$18.219}               \\ \hline
\multirow{2}{*}{\shortstack{BraTS\\T1-CE}}    & Cross-Entropy  & 0.644$\pm$0.142          & 0.489$\pm$0.145          & 20.255$\pm$12.525                        \\
                                              & C$^2$AM (ours) & \textbf{0.695$\pm$0.095} & \textbf{0.540$\pm$0.108} & \textbf{18.129$\pm$12.335}               \\ \hline
\multirow{2}{*}{\shortstack{BraTS\\T2}}       & Cross-Entropy  & 0.684$\pm$0.155          & 0.539$\pm$0.167          & 27.966$\pm$32.604                        \\
                                              & C$^2$AM (ours) & \textbf{0.721$\pm$0.086} & \textbf{0.571$\pm$0.101} & \textbf{14.940$\pm$8.736}                \\ \hline
\multirow{2}{*}{\shortstack{BraTS\\T2-FLAIR}} & Cross-Entropy  & 0.819$\pm$0.105          & 0.705$\pm$0.140          & 12.039$\pm$9.160                         \\
                                              & C$^2$AM (ours) & \textbf{0.862$\pm$0.088} & \textbf{0.767$\pm$0.122} & \textbf{8.664$\pm$6.440}                 \\ \hline
\end{tabular}
\end{table}
\begin{table}
\centering
\caption{Comparison with WSSS Methods using ResNet-50 on BraTS dataset with T1, T1-CE, T2, and T2-FLAIR MRI images. Results are reported in the form of mean$\pm$std. We mark the highest score among WSSS methods with bold text.}
\label{Table:Supple_Res50}
\begin{tabular}{c|c|c|l}
\hline
\multicolumn{4}{c}{BraTS T1} \\ \hline
Method          & Dice $\uparrow$ & IoU $\uparrow$ & \multicolumn{1}{c}{HD95 $\downarrow$} \\ \hline
Grad-CAM (2016) & 0.046$\pm$0.086    & 0.026$\pm$0.053   & 132.469$\pm$33.056                       \\
ScoreCAM (2020) & 0.233$\pm$0.089    & 0.134$\pm$0.058   & 75.245$\pm$15.082                        \\
LFI-CAM (2021)  & 0.166$\pm$0.159    & 0.100$\pm$0.107   & 132.215$\pm$48.849                       \\
LayerCAM (2021) & 0.453$\pm$0.195    & 0.313$\pm$0.164   & 40.190$\pm$55.079                        \\
AME-CAM  (ours) & \textbf{0.587$\pm$0.149}    & \textbf{0.430$\pm$0.141}   & \textbf{20.903$\pm$19.565 }                       \\ \hline
\multicolumn{4}{c}{BraTS T1-CE} \\ \hline
Method          & Dice $\uparrow$ & IoU $\uparrow$ & \multicolumn{1}{c}{HD95 $\downarrow$} \\ \hline
Grad-CAM (2016) & 0.341$\pm$0.132    & 0.213$\pm$0.096   & 48.055$\pm$17.094                        \\
ScoreCAM (2020) & 0.343$\pm$0.235    & 0.234$\pm$0.188   & 50.868$\pm$32.598                        \\
LFI-CAM (2021)  & 0.639$\pm$0.136    & 0.484$\pm$0.143   & \textbf{18.686$\pm$8.465}                         \\
LayerCAM (2021) & 0.480$\pm$0.213    & 0.341$\pm$0.176   & 40.003$\pm$72.395                        \\
AME-CAM  (ours) & \textbf{0.652$\pm$0.099}    & \textbf{0.491$\pm$0.107}   & 19.327$\pm$10.951                        \\ \hline
\multicolumn{4}{c}{BraTS T2} \\ \hline
Method          & Dice $\uparrow$ & IoU $\uparrow$ & \multicolumn{1}{c}{HD95 $\downarrow$} \\ \hline
Grad-CAM (2016) & 0.075$\pm$0.060    & 0.040$\pm$0.035   & 147.272$\pm$26.227                       \\
ScoreCAM (2020) & 0.479$\pm$0.219    & 0.342$\pm$0.193   & 38.509$\pm$27.572                        \\
LFI-CAM (2021)  & 0.523$\pm$0.209    & 0.383$\pm$0.205   & 65.917$\pm$56.581                        \\
LayerCAM (2021) & 0.638$\pm$0.184    & 0.493$\pm$0.190   & 26.907$\pm$33.535                        \\
AME-CAM  (ours) & \textbf{0.682$\pm$0.168}    & \textbf{0.540$\pm$0.175}   & \textbf{19.078$\pm$24.685}                        \\ \hline
\multicolumn{4}{c}{BraTS T2-FLAIR} \\ \hline
Method          & Dice $\uparrow$ & IoU $\uparrow$ & \multicolumn{1}{c}{HD95 $\downarrow$} \\ \hline
Grad-CAM (2016) & 0.221$\pm$0.084    & 0.127$\pm$0.056   & 68.666$\pm$22.125                        \\
ScoreCAM (2020) & 0.588$\pm$0.252    & 0.458$\pm$0.239   & 26.126$\pm$23.682                        \\
LFI-CAM (2021)  & 0.739$\pm$0.164    & 0.611$\pm$0.197   & 14.429$\pm$6.620                         \\
LayerCAM (2021) & 0.638$\pm$0.196    & 0.497$\pm$0.200   & 23.290$\pm$35.905                        \\
AME-CAM  (ours) & \textbf{0.759$\pm$0.132}    & \textbf{0.629$\pm$0.163}   & \textbf{12.696$\pm$8.610}                         \\ \hline
\end{tabular}
\end{table}

% In this section, we show the quantitative comparison among the CAM-based WSSS methods using ResNet-50 as the classifier backbone in Table \ref{Table:Compare_CAM_Res50}. 
% In all modalities of positive patients, the dice scores of LayerCAM[9] are extremely low, which are lower than 0.3.
% As for the ScoreCAM[23] and the LFI-CAM[13], they achieve a significant improvement from LayerCAM with a higher dice score, except for the T1 modality of LFI-CAM. The proposed AME-CAM achieves the optimal dice score in the evaluation of positive patients.
% When it comes to the negative cases, all the previous works still suffer from low dice scores in negative patients, which is due to the false activation issue. Only the proposed method successfully solves the problem and remains a high dice score.
% In the overall evaluation, our AME-CAM achieves the highest dice score and IoU among all the comparisons with other CAM-based approaches. Moreover, the dice score and the IoU are over 0.8 in the T2-FLAIR modality of the BraTS dataset.

% {\small
% \bibliographystyle{splncs04}
% \bibliography{miccai23}
% }